\documentclass{article} 
\usepackage{iclr2023_conference,times}

\usepackage{amsmath,amsfonts,bm}

\def\eqref#1{equation~\ref{#1}}

\def\1{\bm{1}}

\DeclareMathAlphabet{\mathsfit}{\encodingdefault}{\sfdefault}{m}{sl}
\SetMathAlphabet{\mathsfit}{bold}{\encodingdefault}{\sfdefault}{bx}{n}

\usepackage{hyperref}
\usepackage{url}
\usepackage{pifont}
\newcommand{\cmark}{\ding{51}}%
\newcommand{\xmark}{\ding{55}}%
\usepackage{comment}
\usepackage{graphicx}
\usepackage{wrapfig}
\usepackage[linesnumbered,ruled,vlined]{algorithm2e}
\usepackage{multirow}
\newcommand{\share}[1]{\langle#1 \rangle}

\title{Learning to Linearize Deep Neural Networks for Secure and Efficient Private Inference}

\author{Souvik Kundu \thanks{ Part of the work was done when the first author was a graduate student at USC.} \\
Intel Labs, San Diego, USA\\
\texttt{souvikk.kundu@intel.com} \\
\And
Shunlin Lu, Yuke Zhang, Jacqueline T. Liu, \& Peter A. Beerel\\
Department of Electrical and Computer Engineering \\
University of Southern California,
Los Angeles, USA \\
\texttt{\{shunlinlu,yukezhan,jtliu,pabeerel\}@usc.edu} \\
}

\iclrfinalcopy 
\begin{document}

\maketitle

\begin{abstract}
  The large number of ReLU non-linearity operations in existing deep neural networks makes them ill-suited for latency-efficient private inference (PI). Existing techniques to reduce ReLU operations often involve manual effort and sacrifice significant accuracy. In this paper, we first present a novel measure of non-linearity layers' ReLU sensitivity, enabling mitigation of the time-consuming manual efforts in identifying the same. Based on this sensitivity, we then present SENet, a three-stage training method that for a given ReLU budget, automatically assigns per-layer ReLU counts, decides the ReLU locations for each layer's activation map, and trains a model with significantly fewer ReLUs to potentially yield latency and communication efficient PI. Experimental evaluations with multiple models on various datasets show SENet's superior performance both in terms of reduced ReLUs and improved classification accuracy compared to existing alternatives. In particular, SENet can yield models that require up to $\mathord{\sim}2 \times$ fewer ReLUs while yielding similar accuracy. For a similar ReLU budget SENet can yield models with $\mathord{\sim}2.32 \%$ improved classification accuracy, evaluated on CIFAR-100.
\end{abstract}

\section{Introduction}
With the recent proliferation of several AI-driven client-server applications 
including image analysis \citep{litjens2017survey}, object detection, speech recognition \citep{hinton2012deep}, and voice assistance services, the demand for machine learning inference as a service (MLaaS) has grown. 
\begin{wrapfigure}{r}{0.41\textwidth}
\vspace{-4mm}
  \begin{center}
    \includegraphics[width=0.4\textwidth]{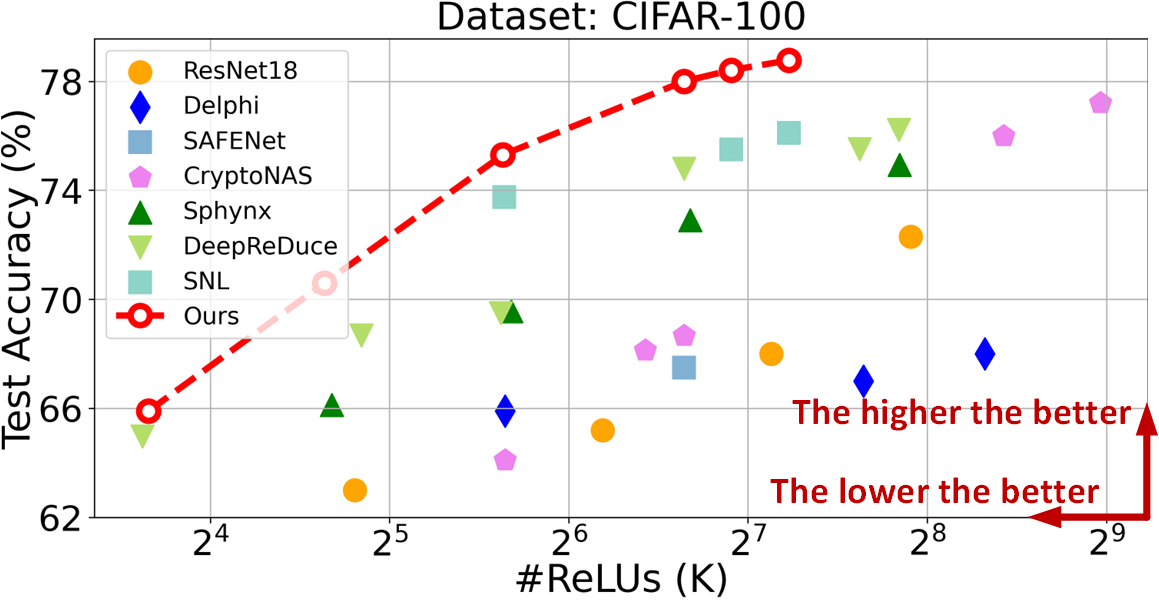}
  \end{center}
  \vspace{-3mm}
  \caption{Comparison of various methods in accuracy vs. \#ReLU trade-off plot. SENet outperforms the existing approaches with an accuracy improvement of up to $\mathord{\sim}4.5\%$ for similar ReLU budget.}
  \vspace{-5mm}
  \label{fig:intro_comparison}
\end{wrapfigure}
Simultaneously, the emergence of privacy concerns from both the users and model developers has made \textit{private inference} (PI) an important aspect of MLaaS. In PI the service provider retains the proprietary models in the cloud where the inference is performed on the client's encrypted data (\textit{ciphertexts}), thus preserving both model \citep{kundu2021analyzing} and data-privacy \citep{yin2020dreaming}. 

Existing PI methods rely on various cryptographic protocols, including homomorphic encryption (HE) \citep{brakerski2014efficient, gentry2009fully} and additive secret sharing (ASS) \citep{goldreich2019play} for the linear operations in the convolutional and fully connected (FC) layers. For example, popular methods like Gazelle \citep{juvekar2018gazelle}, DELPHI \citep{mishra2020delphi}, and Cheetah \citep{reagen2021cheetah} use HE while MiniONN \citep{liu2017oblivious} 
and CryptoNAS \citep{ghodsi2020cryptonas} use ASS. For performing the non-linear ReLU operations, the PI methods generally use Yao's Garbled Circuits (GC) \citep{yao1986generate}. However, GCs demand orders of magnitude higher latency and communication than the PI of linear operations, making latency-efficient PI an exceedingly difficult task. In contrast, standard inference latency is dominated by the linear operations  \citep{kundu2022fast} and is significantly lower than that of PI. 

This has motivated the unique problem of reducing the number of ReLU non-linearity operations to reduce the communication and latency overhead of PI.  In particular, recent literature has leveraged neural architecture search (NAS) to optimize both the number and placement of ReLUs  \citep{ghodsi2020cryptonas, cho2021sphynx}. However, these methods often cost significant accuracy drop, particularly when the ReLU budget is low. For example, with a ReLU budget of 86k, CryptoNAS costs $\mathord{\sim}9\%$ accuracy compared to the model with all ReLUs (AR) present. DeepReDuce \citep{jha2021deepreduce} used a careful multi-stage optimization and provided reduced accuracy drop of $\mathord{\sim}3\%$  at similar ReLU budgets.  However, DeepReDuce heavily relies on manual effort for the precise removal of ReLU layers, making this strategy exceedingly difficult, particularly, for models with many layers. A portion of these accuracy drops can be attributed to the fact that these approaches are constrained to remove ReLUs at a higher granularity of layers and channels rather than at the pixel level. Only very recently, \citep{cho2022selective} proposed $l_1$-regularized  pixel level ReLU reduction. However, such approaches are extremely hyperparameter sensitive and often \textcolor{black}{do} not guarantee meeting a specific ReLU budget. Moreover, the large number of training iterations required for improved accuracy may not be suitable for compute-limited servers \citep{mishra2020delphi}. 

\textbf{Our contributions.} Our contribution is three-fold. We first empirically demonstrate the relation between a layer's sensitivity towards pruning and its associated ReLU sensitivity. 
Based on our observations, we introduce an automated layer-wise ReLU sensitivity evaluation strategy and propose SENet, a three-stage training process to yield secure and efficient networks for PI that guarantees meeting a target ReLU budget without any hyperparameter-dependent iterative training. In particular, for a given global ReLU budget, we first determine a sensitivity-driven layer-wise non-linearity (ReLU) unit budget. Given this budget, we then present a layer-wise ReLU allocation mask search. For each layer, we evaluate a binary mask tensor with the size of the corresponding activation map for which a 1 or 0 signifies the presence or absence of a ReLU unit, respectively. Finally, we use the trained mask to create a partial ReLU (PR) model with ReLU present only at fixed parts of the non-linearity layers, and fine-tune it via distillation from an iso-architecture trained AR model. \textcolor{black}{Importantly, we support ReLU mask allocation both at the granularity of individual pixels and activation channels.} 

To further reduce both linear and non-linear (ReLU) layer compute costs, we extend our approach to SENet++. SENet++ uses a single training loop to train a model of different channel dropout rates (DRs) $d_r$ ($d_r \le 1.0$) of the weight tensor, where each $d_r$  yields a sub-model with a MAC-ReLU budget smaller than or same as that of the original one. In particular, inspired by the idea of ordered dropout \citep{horvath2021fjord}, we train a PR model with multiple dropout rates \citep{horvath2021fjord}, where each dropout rate corresponds to a scaled channel sub-model having number of channels per layer $\propto$ the $d_r$. This essentially allows the server to yield multiple sub-models for different compute requirements that too via a single training loop, without costly memory footprint. 
Table \ref{table:motive1_trans} compares the important characteristics of our methods with existing alternatives.

We conduct extensive experiments and ablations on various models including variants of ResNet, Wide Residual Networks, and VGG  on CIFAR-10, CIFAR-100, Tiny-ImageNet, \textcolor{black}{and ImageNet} datasets. Experimental results show that SENet can yield SOTA accuracy-ReLU trade-off with an improved accuracy of up to $\mathord{\sim}2.32\%$ for similar ReLU budgets. SENet++ ($d_r = 0.5$) can further improve the MAC and ReLU cost of SENet,  with an additional saving of $4\times$ and $\mathord{\sim}2\times$, respectively.

\begin{table}[!t]
	\tiny\addtolength{\tabcolsep}{-6.5pt}
		\caption{\textcolor{black}{Comparison between existing approaches in yielding efficient models  to perform PI. Note, SENet++ can yield a model that can be switched to sub-models of reduced channel sizes.}}
		\label{table:motive1_trans}
		\centering
		\begin{tabular}{c|c|c|c|c|c}
			\hline
			\textbf{Name} & \textbf{Method} & \textbf{Reduced}  & \textbf{Granularity} & \textbf{Reduce model} & \textbf{Supports dynamic} \\
			 & \textbf{used} & \textbf{non-linearity} &  &  \textbf{dimension} & \textbf{channel dropping} \\
			\hline
			Irregular pruning  & Various  & \textcolor{red}{\xmark} & Scalar weight &  \textcolor{red}{\xmark} &  \textcolor{red}{\xmark} \\   
			\hline
			Structured pruning  & Various  & \textcolor{green}{\cmark} & Channel, filter &  \textcolor{green}{\cmark} &  \textcolor{red}{\xmark}\\   
			\hline
			Sphynx \citep{cho2021sphynx}  & NAS  & \textcolor{green}{\cmark} & Layer-block & \textcolor{red}{\xmark} &  \textcolor{red}{\xmark}\\ 
			\hline
			CryptoNAS \citep{ghodsi2020cryptonas} & NAS  & \textcolor{green}{\cmark} & Layer-block & \textcolor{red}{\xmark} &  \textcolor{red}{\xmark}\\
			\hline
			DELPHI \citep{mishra2020delphi} & NAS + PA  & \textcolor{green}{\cmark} & Layer-block & \textcolor{red}{\xmark} &  \textcolor{red}{\xmark}\\
			\hline
			SAFENet \citep{lou2020safenet} & NAS + PA & \textcolor{green}{\cmark} & Channel & \textcolor{red}{\xmark} &  \textcolor{red}{\xmark}\\
			\hline
			DeepReDuce \citep{jha2021deepreduce} & Manual + HE  & \textcolor{green}{\cmark} & Layer-block & \textcolor{red}{\xmark} &  \textcolor{red}{\xmark}\\
			\hline
			\textcolor{black}{SNL \citep{cho2022selective}} & \textcolor{black}{$l_1$-regularized} & \textcolor{green}{\cmark} & \textcolor{black}{Channel, pixel} & \textcolor{red}{\xmark} & \textcolor{red}{\xmark} \\
			\hline
			SENet (ours) & Automated  & \textcolor{green}{\cmark} & \textcolor{black}{Channel, pixel} & \textcolor{red}{\xmark} & \textcolor{red}{\xmark}\\
			\hline
			SENet++ (ours) & Automated  & \textcolor{green}{\cmark} & \textcolor{black}{Channel, pixel} & \textcolor{green}{\cmark} & \textcolor{green}{\cmark}\\
			\hline
		\end{tabular}
		\vspace{-6mm}
\end{table} 

\section{Preliminaries and Related Work}
\textbf{Cryptographic primitives.} We briefly describe the relevant cryptographic primitives in this section. 

\textit{Additive secret sharing.} Given an element $x$, an ASS of $x$ is the pair $(\share{x}_1, \share{x}_2)=(x-r, r)$, where $r$ is a random element and $x=\share{x}_1+\share{x}_2$. Since $r$ is random, the value $x$ cannot be revealed by a single share, so that the value $x$ is hidden.

\textit{Homomorphic encryption.} HE ~\citep{gentry2009fully} is a public key encryption scheme that supports homomorphic operations on the ciphertexts. Here, encryption function $E$ generates the ciphertext $t$ of a plaintext $m$ where $t=E(m, pk)$, and a decryption function $D$ obtains the plaintext $m$ via $m=D(t, sk)$, where $pk$ and $sk$ are corresponding public and secret key, respectively. 
In PI, the results of linear operations can be obtained homomorphically through $m_1 \circ m_2 = D(t_1  \star t_2, sk)$, where $\circ$ represents a linear operation, $\star$ is its corresponding homomorphic operation, $t_1$ and $t_2$ are the ciphertexts of $m_1$ and $m_2$, respectively.

\textit{Garbled circuits.} GC~\citep{yao1986generate} allows two parties to jointly compute a Boolean function $f$ over their private inputs without revealing their inputs to each other. The Boolean function $f$ is represented as a Boolean circuit $C$. Here, a garbler creates an encoded Boolean circuit $\tilde{C}$ and a set of input-correspondent labels through a procedure $Garble(C)$ to send $\tilde{C}$ and the labels to the other party who acts as an evaluator. The evaluator further sends the output label upon evaluation via  $Eval(\tilde{C})$. Finally, the garbler decrypts the labels to get the plain results to share with the evaluator. 

\textbf{Private inference.} Similar to \citep{mishra2020delphi}, in this paper, we focus on a semi-honest client-server PI scenario where a client, holding private data, intends to use inference service from a server having a private model. 
Specifically, the semi-honest parties strictly follow the protocol but try to reveal their collaborator's private data by inspecting the information they received. On the other hand, a malicious client could deviate from the protocol. 

To defend against various threats existing cryptographic protocols \citep{mishra2020delphi, ghodsi2020cryptonas, lou2020safenet}  rely on the popular online-offline topology \citep{mishra2020delphi}, where the client data independent component is pre-computed in the offline phase \citep{juvekar2018gazelle, mishra2020delphi, zahra2021circa, ryan2021muse}. For the linear operations, DELPHI~\citep{mishra2020delphi} and MiniONN~\citep{liu2017oblivious} move the heavy primitives in HE and \textcolor{black}{ASS} to offline enabling fast linear operations during PI online stage. However, the compute-heavy $Eval(\tilde{C})$ of GC keeps the ReLU cost high even at the online stage. 

\textbf{ReLU reduction for efficient PI.} Existing works use model designing with reduced ReLU counts via either search for efficient models \citep{mishra2020delphi, ghodsi2020cryptonas, lou2020safenet, cho2021sphynx} or manual re-design from an existing model \citep{jha2021deepreduce}. In particular, SAFENet~\citep{lou2020safenet} enables more fine-grained channel-wise substitution and mixed-precision activation approximation. CryptoNAS~\citep{ghodsi2020cryptonas} re-designs the neural architectures through evolutionary NAS techniques to minimize ReLU operations. Sphynx~\citep{cho2021sphynx} further improves the search by leveraging differentiable macro-search NAS \citep{liu2018darts} in yielding efficient PI models. DeepReDuce \citep{jha2021deepreduce}, on the other hand, reduced ReLU models via a manual effort of finding and dropping redundant ReLU layers starting from  an existing standard model. Finally, a recent work \citep{cho2022selective} leveraged $l_1$-regularization to remove ReLU at the pixel level to yield SOTA accuracy vs non-linearity trade-off. However, the extreme hyperparameter dependence of such methods often provide sub-optimal solution and does not necessarily guarantee meeting a target ReLU budget. Moreover, a resource-limited server \citep{mishra2020delphi} may not afford costly iterative training \citep{cho2022selective} in reducing the ReLU count.



\section{Motivational Study: Relation between ReLU importance and Pruning Sensitivity}
\label{sec:motivation}

Existing work to find the importance of a ReLU layer \citep{jha2021deepreduce}, requires manual effort and is extremely time consuming. In contrast, model pruning literature \citep{lee2018snip, kundu2021dnr} leveraged various metrics to efficiently identify a layer's sensitivity towards a target pruning ratio. In particular, a layer's \textit{pruning sensitivity} can be quantitatively defined as the accuracy reduction caused by pruning a certain ratio of parameters from it \citep{ding2019global}. 
In particular, recent literature leveraged sparse learning \citep{ding2019global, kundu2021dnr} and used a trained sparse model to evaluate the sensitivity of a layer $l$ ($\eta_{\bm{\theta}^l}$) as the ratio $\frac{\texttt{total \# of non-zero layer parameters}}{\texttt{total \# layer parameters}}$. \textbf{Despite significant progress in weight pruning sensitivity, due to the absence of any trainable parameter in the ReLU layer, its sensitivity for a given ReLU budget is yet to be explored}. We hypothesize that there may be a correlation between a layer's pruning sensitivity \citep{kundu2021dnr} and the importance of ReLU and have conducted the following experiments to explore this.

Let us assume an $L$-layer DNN model $\Phi$ parameterized by $\bm{\Theta} \in \mathbb{R}^m$ that learns a function $f_{\Phi}$, where $m$ represents the total number of model parameters.

The goal of DNN parameter pruning is to identify and remove the unimportant parameters from a DNN and yield a reduced parameter model that has comparable performance to the baseline unpruned model. As part of the pruning process for a given parameter density $d$, each parameter is associated with an auxiliary indicator variable $c$ belonging to a mask tensor $\bm{c} \in \{0,1\}^m$ such that only those $\theta$ remain non-zero whose corresponding $c = 1$. With these notations, we can formulate the training optimization as
\vspace{-2mm}
\begin{align}
     \text{min  } \mathcal{L}(f_{\Phi}(\bm{\Theta} \odot \bm{c})) \text{,  s.t. } ||\bm{c}||_0 \le d \times m
     \label{eq:loss_ce_mask}
\end{align}
\noindent
\textcolor{black}{where $\mathcal{L}(.)$ represents the loss function, which for image classification tasks is generally the cross-entropy (CE) loss. We used a sparse learning framework  \citep{kundu2021dnr} to train}
\begin{wrapfigure}{r}{0.40\textwidth}
\vspace{-4mm}
  \begin{center}
    \includegraphics[width=0.40\textwidth]{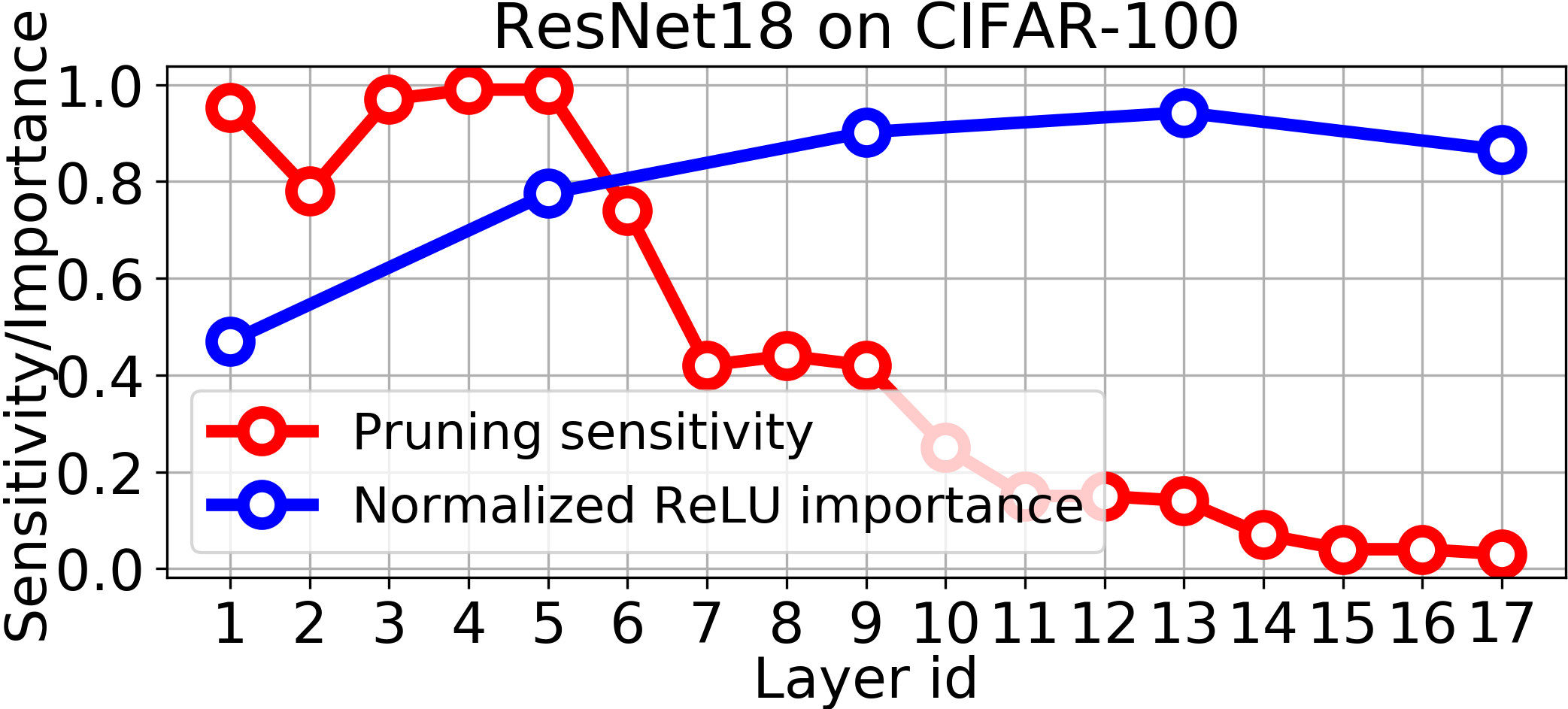}
  \end{center}
  \vspace{-4mm}
  \caption{Layer-wise pruning sensitivity ($d=0.1$) vs. normalized ReLU importance. The later layers are less sensitive to pruning, and, thus, can afford significantly more zero-valued weights as opposed to the earlier ones. On the contrary,  later ReLU stages generally have higher importance.}
  \vspace{-4mm}
  \label{fig:relu_utility_prune_sens}
\end{wrapfigure}
      a ResNet18 on CIFAR-100 for a target $d=0.1$ and computed the pruning sensitivity of each layer. In particular, as shown in Fig. \ref{fig:relu_utility_prune_sens}, earlier layers have higher pruning sensitivity than later ones. This means that to achieve close to baseline performance, the model trains later layers' parameters towards zero more than those of earlier layers. 

We then compared this trend with that of the importance of different ReLU layers as defined in \citep{jha2021deepreduce}. \textcolor{black}{   In particular, we first identified five different modules of ReLU placement in a ResNet18, the pre-basic-block (BB) stem, BB1, BB2, BB3, and BB4.  We then created five ResNet18 variants with ReLU non-linearity present only at one of the modules while replacing non-linearity at the other modules with identity layers.} We identify the modules yielding higher accuracy to be the ones with higher ReLU importance \citep{jha2021deepreduce}. We then normalized the importance of a ReLU  stage with accuracy $\texttt{Acc}$ as the ratio ${(\texttt{Acc} - \texttt{Acc}_{min})}/{(\texttt{Acc}_{max} - \texttt{Acc}_{min})}$. Here $\texttt{Acc}_{max}$ and $\texttt{Acc}_{min}$ correspond to the accuracy of models with all and no ReLUs, respectively.

As depicted in Figure \ref{fig:relu_utility_prune_sens}, the results show that the ReLU importance and parameter pruning sensitivity of a layer are inversely correlated.  This inverse correlation may
imply that a pruned layer can afford to have more zero-valued weights when the associated ReLU layer forces most of the computed activation values to zero.

\section{SENet Training Methodology}

As highlighted earlier, for a large number of ReLU layers $L_r$, the manual evaluation and analysis of the candidate architectures become inefficient and time consuming. Moreover, the manual assignment of ReLU at the pixel level becomes even more intractable because the number of pixels that must be considered, explodes. To that end, we now present SENet, a three-stage automated ReLU trimming strategy that can yield models for a given reduced ReLU budget.

\begin{algorithm}[!t]
\scriptsize
\SetAlgoLined
\DontPrintSemicolon
\KwData{Global ReLU budget $r$, model parameters $\bm{\Theta}$, model parameter proxy density $d$,} number of ReLU layers $L_r$, active ReLU indicator $\bm{a} \in \{1\}^{L_r}$\\
\textbf{Output:} Per-layer \# ReLU count.\\
$ \bm{\eta}_{\bm{\alpha}} \leftarrow \texttt{evalActSens(}\bm{\Theta}, d \text{)} $\;
\For{$\text{l} \leftarrow 0$ \KwTo {$L_r$}}
{
    $\eta_{\bm{\alpha}^l} \leftarrow \frac{\eta_{\bm{\alpha}^l}}{\sum^L_{i=0} \eta_{\bm{\alpha}^i} \times a^i}$\;
}
$\texttt{initVals(}r_{remain}, r_{total}, \bm{r}_{final})$\;
\While{$r_{total} <  r$}
{   
    \For{$\text{l} \leftarrow 0$ \KwTo {L}}
    {
    $r^l_{cur} \leftarrow \texttt{assignReluProportion(}r_{remain}, \eta_{\bm{\alpha}^l}, \bm{a})$\;
    $r^l_{final}, r_{total} \leftarrow \texttt{assignUpdateRelu(}r^l_{final}, r^l_{cur}, r_{total})$\;
    }
    
}
$r_{remove} \leftarrow r_{total} - r$\;
\While{$r_{remove} >  0$}
{   
    \For{$\text{l} \leftarrow 0$ \KwTo {L}}
    {
    $r^l_{cur} \leftarrow \texttt{removeReluProportion(}r_{del}, \eta_{\bm{\alpha}^l}, \bm{a})$\;
    $r^l_{final}, r_{remove} \leftarrow \texttt{removeUpdateRelu(}r^l_{final}, r^l_{cur}, r_{remove})$\;
    }
    
}
$\texttt{return } \bm{r}_{final}$\;
\vspace{-1mm}
 \caption{Layer-wise \#ReLU Allocation Algorithm}
 \label{alg:relu_alloc}
\end{algorithm}

\subsection{Sensitivity Analysis}
\label{subsec:sensitivity_analysis}



Inspired by our observations in Section \ref{sec:motivation}, we define the ReLU sensitivity of a layer $l$ as
\begin{align}
    \eta_{\bm{\alpha}^l} = (1 - \eta_{\bm{\theta}^l})
\end{align}
\noindent
It is important to emphasize that, unlike ReLU importance, ReLU sensitivity does not require training many candidate models. However, $\eta_{\bm{\theta}^l}$ can only be evaluated for a specific $d$. We empirically observe that  $d > 0.3$ tends to yield uniform sensitivity across layers due to a large parameter budget. In contrast, ultra-low density $d < 0.1$, costs non-negligible accuracy drops \citep{liu2018rethinking, kundu2022towards}. Based on these observations, we propose to quantify ReLU sensitivity with a {\em proxy density} of $d = 0.1$. 

Moreover, to avoid the compute-heavy pruning process, we leverage the idea of sensitivity evaluation before training \citep{lee2018snip}. On a sampled mini batch from training data $\mathcal{D}$, the sensitivity of the $j^{th}$ connection with associated indication variable and vector as $c_j$ and $\bm{e}_j$, can be evaluated as,
\vspace{-0mm}
\begin{align}
    \Delta{\mathcal{L}}_j(f_{\Phi}(\bm{\Theta}; \mathcal{\bm{D}})) &= g_j(f_{\Phi}(\bm{\Theta}; \mathcal{\bm{D}})) = \frac{\partial{\mathcal{L}(f_{\Phi}(\bm{c} \odot \bm{\Theta}; \mathcal{\bm{D}}))}}{\partial{c_j}}\Bigr\rvert_{\bm{c}=\bm{1}} \\ \nonumber
    &= \lim_{\delta \rightarrow 0} \frac{\mathcal{L}(f_{\Phi}(\bm{c} \odot \bm{\Theta}; \mathcal{\bm{D}})) - \mathcal{L}(f_{\Phi}((\bm{c} - \delta \bm{e}_j) \odot \bm{\Theta}; \mathcal{\bm{D}}))}{\delta} \Bigr\rvert_{\bm{c}=\bm{1}}
\end{align}
\vspace{-0mm}

\noindent
 \textcolor{black}{where $\bm{c}$ is a vector containing all indicator variables. The $\frac{\partial \mathcal{L}}{\partial c_j}$ is an infinitesimal version of $\Delta{\mathcal{L}}_j$ \textcolor{black}{measuring the impact of a change in $c_j$ from $1 \rightarrow 1 - \delta$}. It can be computed using one forward pass for all $j$ at once.} We normalize the connection sensitivities, rank them, and identify the top d-fraction of connections. We then define the layer sensitivity $\eta_{\bm{\Theta}^l}$ as the fraction of connections of each layer that are in the top \textcolor{black}{d-fraction}. For a given global ReLU budget $r$, we then assign the $\#$ ReLU for each layer proportional to its normalized ReLU sensitivity.  The details are shown in Algorithm \ref{alg:relu_alloc} (Fig. \ref{fig:proposed_framework} as point \textcircled{1}). \textcolor{black}{Note $r^l_{final}$ in Algorithm \ref{alg:relu_alloc} represents the allocated \#ReLUs of layer $l$ at the end of stage 1, with $\bm{r}_{final}$ representing the set of \#ReLUs for all the ReLU  layers.}

\subsection{ReLU Mask Identification}
\label{subsec:relu_mask_eval}
After layer-wise \#ReLU allocation, we identify the ReLU locations in each layer's activation map. \textcolor{black}{In particular}, for a non-linear layer $l$, we assign a mask tensor $M^l \in \{0,1\}^{h^l\times w^l \times c^l}$, where $h^l, w^l$, and $c^l$ represents the height, width, and the number of channels in the activation map. For a layer $l$, we initialize $M$ with $r^l_{final}$ assigned 1's with random locations. 
Then we perform a distillation-based training of the PR model performing ReLU ops only at the locations of the masks with 1, while distilling knowledge from an AR model of the same architecture (see Fig. \ref{fig:proposed_framework}, point \textcircled{2}). At the end of each epoch, for each layer $l$, we rank the top-$r^l_{final}$ locations based on the highest absolute difference between the PR and AR model's post-ReLU activation output (averaged over all the mini-batches) for that layer, and update the $M^l$ with 1's at these locations. This, on average, de-emphasizes the locations where the post-ReLU activations in both the PR and AR models are positive.
We terminate mask evaluation once the ReLU mask\footnote{The identified mask tensor has non-zeros irregularly placed. This can be easily extended to the generation of the structured mask, by allowing the assignment and removal of mask values at the granularity of channels instead of activation scalar \citep{kundu2021dnr}.} evaluation reaches the maximum mask training epochs or when the normalized hamming distance between masks generated after two consecutive epochs is below a certain pre-defined $\epsilon$ value. \textbf{Notably, there has been significant research in identifying important trainable parameters \citep{savarese2020winning,kusupati2020soft, kundu2020pre, kundu2022bmpq,kundu2022fast,babakniya2022federated} through various proxies including magnitude, gradient, Hessian, however, due to the absence of any trainable parameter in the ReLU layer, such methods can't be deployed in identifying important ReLU units of a layer.} 

\textcolor{black}{\textbf{Channel-wise ReLU mask identification.} The mask identification technique described above, creates irregular ReLU masks. To support a coarser level of granularity where the ReLU removal happens "channel-wise", we now present a simple yet effective extension of the mask identification. For a layer $l$, we first translate the total non-zero ReLU counts to total non-zero ReLU channels as $r^l_c = \lceil \frac{r^l_{final}}{h^lw^l} \rceil$. We then follow the same procedure as irregular mask identification, however, only keep top-$r^l_c$ channels as non-zero.}
 
\subsection{Maximizing Activation Similarity via Distillation}

\label{subsec:post_relu_kd}
\begin{figure}[!t]
\includegraphics[width=0.86\textwidth]{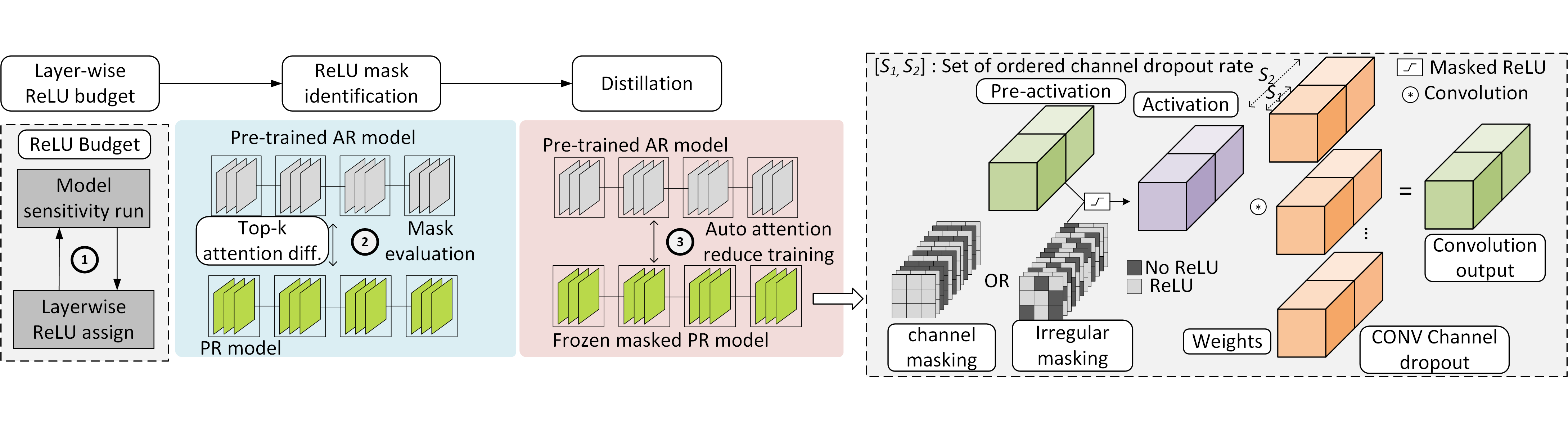}
\centering
\vspace{-6mm}
   \caption{\textcolor{black}{Different stages of the proposed training methodology for efficient private inference with dynamic channel reduction. 
   For example, the model here supports two channel SFs, $S_1$ and $S_2$. Note, similar to \citep{horvath2021fjord}, for each SF support we use a separate batch-normalization (BN) layer to maintain separate statistics.}}
\label{fig:proposed_framework}
\vspace{-5mm}
\end{figure}

Once the mask for each layer is frozen, we start our final training phase in which we maximize the similarity between activation functions of our PR and AR models, see Fig. \ref{fig:proposed_framework}, point \textcircled{3}. In particular, we initialize a PR model with the weights and mask of best PR model of stage 2 and allow only the parameters to train. We train the PR model with distillation via KL-divergence loss \citep{hinton2015distilling,kundu2021attentionlite} from a pre-trained AR  along with a CE-loss.  Moreover, we introduce an AR-PR post-ReLU activation mismatch (PRAM) penalty into the loss function. This loss drives the PR model to have activation maps that are similar to that of the AR model.

More formally, let $\Psi_{pr}^m$ and $\Psi_{ar}^m$ represent the $m^{th}$ pair of vectorized post-ReLU activation maps of same layer for $\Phi_{pr}$ and $\Phi_{ar}$, respectively. Our loss function for the fine-tuning phase is given as
\begin{align}\label{eq:sp_distill_loss}
        {\mathcal{L}} = (1-\lambda)\underbrace{{\mathcal{L}}_{pr}(y,y^{pr})}_{\text{CE loss}} + & \lambda\underbrace{{\mathcal{L}}_{KL}\left(\sigma\left(\frac{z^{ar}}{\rho}\right),\sigma\left(\frac{z^{pr}}{\rho}\right)\right)}_{\text{KL-div. loss}}
    + \frac{\beta}{2}\sum_{m \in I}\underbrace{\left\lVert\frac{\Psi_{pr}^m}{\lVert \Psi_{pr}^m\rVert_2} - \frac{\Psi_{ar}^m}{\lVert \Psi_{ar}^m\rVert_2}\right\lVert_2}_{\text{PRAM loss}}
\end{align}
\noindent
where $\sigma$ represents the softmax function with $\rho$ being its temperature. $\lambda$ balances the importance between the CE and KL divergence loss components, and $\beta$ is the weight for the PRAM loss. Similar to \citep{zagoruyko2016paying}, we use the $l_2$-norm of the normalized activation maps to compute this loss.

\subsection{SENet++: Support for Ordered Channel Dropping}

To yield further compute-communication benefits, we now present an extension of SENet, namely SENet++, that can perform the ReLU reduction while also supporting inference with reduced model sizes. In particular, we leverage the idea of ordered dropout (OD) \citep{horvath2021fjord} to simultaneously train multiple sub-models with different fractions of channels. The OD method is parameterized by a candidate dropout set $\mathcal{D}_r$ with dropout rate values $d_r \in (0, 1]$. At a selected $d_r$ for any layer $l$, the model uses a $d_r$-sub-model with only the channels with indices $\{0,1,..., \lceil d_r \cdot C_l \rceil-1\}$ active, effectively pruning the remaining $\{\lceil d_r \cdot C_l \rceil,...,C_l-1\}$ channels. Hence, during training, the selection of a $d_r$-sub-model with $d_r < 1.0 \in \mathcal{D}_r$, is a form of channel pruning, while $d_r = 1.0$ trains the full model.  
For each mini-batch of data, we perform a forward pass once for each value of $d_r$ in $\mathcal{D}_r$, accumulating the loss. We then perform a backward pass in which the model parameters are updated based on the gradients computed on the accumulated loss. 
We first train an AR model with a dropout set $\mathcal{D}_r$. For the ReLU budget evaluation, we consider only the model with $d_r = 1.0$, and finalize the mask by following the methods in Sections \ref{subsec:sensitivity_analysis} and \ref{subsec:relu_mask_eval}. During the maximizing of activation similarity stage, we fine-tune the PR model supporting the same set $\mathcal{D}_r$ as that of the AR model. In particular, the loss function for the fine-tuning is the same as \ref{eq:sp_distill_loss}, for $d_r = 1.0$. For $d_r < 1.0$, we exclude the PRAM loss because we empirically observed that adding the PRAM loss for each sub-model on average does not  improve accuracy. During inference, SENet++ models can be dynamically switched to support reduced channel widths, reducing the number of both ReLUs and MACs compared to the baseline model. 

\section{Experiments}
%
%
\subsection{Experimental Setup}
\textbf{Models and Datasets.} To evaluate the efficacy of the SENet yielded models, we performed extensive experiments on three popular datasets, CIFAR-10, CIFAR-100 \citep{krizhevsky2009learning}, Tiny-ImageNet \citep{hansen2015tiny}, \textcolor{black}{and  ImageNet\footnote{\textcolor{black}{On ImageNet, for comprehensive training with limited resources, we sample 100 classes from the ImageNet dataset with 500 and 50 training and test examples per class, respectively.}}} with three different model variants, namely ResNet (ResNet18, ResNet34) \citep{he2016deep}, wide residual network (WRN22-8) \citep{zagoruyko2016wide}, and VGG (VGG16) \citep{simonyan2014very}. We used PyTorch API to define and train our models on an Nvidia RTX 2080 Ti GPU.

\textbf{Training Hyperparameters.} \textcolor{black}{We performed standard data augmentation (horizontal flip and random} 
\begin{wraptable}{r}{0.43\textwidth}
\vspace{-6mm}
	\tiny\addtolength{\tabcolsep}{-4.0pt}
		\caption{\textcolor{black}{Runtime and communication costs of linear and ReLU operations for $15$-bit fixed-point model parameters/inputs and $31$-bit ReLUs \citep{mishra2020delphi}}.}
		\label{table:cost_report}
		\centering
		\begin{tabular}{c|c|c|c}
			\hline
			 \textbf{Operation} & \textbf{Mode} & \textbf{Runtime}($\mu s$)  & \textbf{Comm. cost}(KB) \\
			 \hline
			 Linear & Offline & 32.6 & 0.095 \\
			  & Online & 0.248 & 0.000563 \\
			 \hline
			 ReLU & Offline & 154.9 & 17.5 \\
			  & Online & 85.3 & 2.048 \\
			  \hline
		\end{tabular}
		\vspace{-3.0mm}
\end{wraptable} 
 cropping with reflective padding) and the SGD optimizer for all training. We trained the baseline all-ReLU model for \textcolor{black}{240, 120, and 60 epochs for CIFAR, Tiny-ImageNet, and ImageNet respectively}, with a starting learning rate (LR) of 0.05 that decays by a factor of 0.1 at the 62.5\%, 75\%, and 87.5\% training epochs completion points. For all the training we used an weight decay coefficient of $5\times 10^{-4}$. For a target ReLU budget, we performed the mask evaluation for \textcolor{black}{150, 100, and 30 epochs, respectively, for the three dataset types} with the $\epsilon$ set to $0.05$, meaning the training prematurely terminates when less than $5\%$ of the total \#ReLU masks change their positions. Finally, we performed the post-ReLU activation similarity improvement for \textcolor{black}{180, 120, and 50 epochs, for CIFAR, Tiny-ImageNet, and ImageNet respectively}. 
Also, unless stated otherwise, we use $\lambda = 0.9$, and $\beta = 1000$ for the loss described in Eq. \ref{eq:sp_distill_loss}. Further details of our training hyper-parameter choices are provided in the Appendix. In Table \ref{table:comparison_perf_table}, we report the accuracy averaged over three runs.
\subsection{SENet Results}
As shown in Table \ref{table:comparison_perf_table}, SENet yields models that have higher accuracy than existing alternatives by
\begin{wraptable}{r}{0.53\textwidth}
\vspace{-4mm}
	\tiny\addtolength{\tabcolsep}{-5.3pt}
		\caption{\textcolor{black}{Results on Tiny-ImageNet and ImageNet.}}
		\label{table:comparison_perf_table_tiny_iamgenet}
		\centering
		\begin{tabular}{c|c|c|c|c|c|c}
			\hline
			 \textbf{Model} & \textbf{Baseline} & \#\textbf{ReLU}  & \textbf{Method} & \textbf{Test} & \textbf{Acc}\%/ & \textbf{Comm.}\\
			 & \textbf{Acc}\% &  \textbf{(k)}  &  &\textbf{Acc}\% & \#\textbf{1k ReLU} &  \textbf{Savings} \\
			 \hline
			 \multicolumn{7}{c}{\textcolor{black}{Dataset: Tiny-ImageNet}}\\
			\hline
			  &  & $142$ & SENet & $58.9$ & $0.414$ & $\times 15.7$ \\
			 ResNet18 & $66.1$  &   $298$ &  &  $\bm {64.96}$ & $0.218$ & $ 7.5\times$ \\
			 \cline{3-7}
			 &   &   $393$ & DeepReDuce\citep{jha2021deepreduce} &  $61.65$ & $0.157$ & $5.7\times $\\
			 &   &   $917$ &  &  $64.66$ & $0.071$ & $\times 2.4$ \\
			\hline
			\multicolumn{7}{c}{\textcolor{black}{Dataset: ImageNet}}\\
			\hline
			\textcolor{black}{ResNet18} & \textcolor{black}{$71.94$}  &   \textcolor{black}{$600$} & \textcolor{black}{SENet} &  \textcolor{black}{$70.28$} & \textcolor{black}{$0.117$} & \textcolor{black}{$3.86\times $} \\
			 &  &   \textcolor{black}{$950$} &  &  \textcolor{black}{$\bm{71.16}$} & \textcolor{black}{$0.075$} & \textcolor{black}{$ 2.43\times$} \\
			 \hline
		\end{tabular}
		\vspace{-1mm}
	\tiny\addtolength{\tabcolsep}{-0.0pt}
		\caption{Results with ReLU reduction at the granularity of  activation channel evaluated on CIFAR-100.}
		\label{table:channel_act_relu}
		\centering
		\begin{tabular}{c|c|c|c|c|c|c}
			\hline
			 \textbf{Model} & \textbf{Baseline} & \#\textbf{ReLU} & \textbf{Method} & \textbf{Test} & \textbf{Acc}\%/ & \textbf{Comm.}\\
			 & \textbf{Acc}\% &  \textbf{(k)} &  &\textbf{Acc}\% & \#\textbf{1k ReLU} &  \textbf{Savings} \\
			 \hline
			  &  & $\bm{180}$ & SENet & $79.02$ & $0.44$ & $\bm{7.7}\times $ \\
			  WRN22-8 & $80.82$ &   $240$  & SENet & $\bm {79.3}$ & $0.33$ & $5.8\times$ \\
			  & &   $200$  & SNL \citep{cho2022selective} & $77.45$ & $0.38$ & $6.9\times$ \\
			\hline
		\end{tabular}
		\vspace{-3mm}
\end{wraptable}
 a significant margin while often requiring fewer ReLUs. For example, at a small  ReLU  budget of $\le 100$k,  our models yield up to $4.15\%$ and $7.8\%$ higher accuracy, on CIFAR-10 and CIFAR-100, respectively. At a ReLU budget of $\le 500$k, our improvement is up to $0.50\%$ and $2.38\%$, respectively, on the two datasets. We further evaluate the communication saving due to the non-linearity reduction by taking the  per ReLU communication cost mentioned in Table \ref{table:cost_report}. In particular, the communication saving reported in the $8^{th}$ column of Table \ref{table:comparison_perf_table} is computed as the ratio of communication costs associated with an AR model to that of the corresponding PR model with reduced ReLUs. We did not report any saving for the custom models, as they do not have any corresponding AR baseline model. On Tiny-ImageNet, SENet models can provide up to $0.3\%$ higher performance while requiring $3.08\times$ fewer ReLUs (Table \ref{table:comparison_perf_table_tiny_iamgenet}). \textcolor{black}{More importantly, even for a high resolution dataset like ImageNet, SENet models can yield close to the baseline performance, depicting the efficacy of our proposed training method.}
\begin{table}[!t]
	\tiny\addtolength{\tabcolsep}{-2.5pt}
		\caption{Performance of SENet and other methods on various datasets and models.}
		\label{table:comparison_perf_table}
		\centering
		\begin{tabular}{c|c|c|c|c|c|c|c}
			\hline
			\textbf{min} $\le$ \textbf{r} $\le$ \textbf{max} & \textbf{Model} & \textbf{Baseline} & \#\textbf{ReLU (k)}  & \textbf{Method} & \textbf{Test} & \textbf{Acc}\%/ & \textbf{Comm.}\\
			  &  & \textbf{Acc}\% &    &  &\textbf{Acc}\% & \#\textbf{1k ReLU} &  \textbf{Savings} \\
			 \hline
			 \hline
			 \multicolumn{8}{c}{Dataset: CIFAR-10}\\
			\hline
			& VGG16 & $93.8$ & $12.5$ &   & $91.6$ & $7.33$ & $23.6\times$\\
			 &  &  & $49.2$ & SENet(ours)  & $93.16$ & $1.89$ & $6.0\times$\\
			\cline{2-4} \cline{6-8}
			$ 0 \le r \le 100 $k  & ResNet18 & $95.2$ & $49.1$ &   & $93.60$ & $1.9$ &$11.3\times$ \\
			&  &  & $82$ &  & $\bm{93.05}$ & $1.14$ & $6.8\times$ \\
			\cline{2-8}
			 & \textcolor{black}{ResNet18} & \textcolor{black}{$95.2$} & \textcolor{black}{$12.9$} &  \textcolor{black}{SNL} \citep{cho2022selective} & \textcolor{black}{$88.23$} & \textcolor{black}{$6.84$} & \textcolor{black}{$43.1\times$}\\
			\cline{2-8}
			 & VGG16 & $93.8$ & $36.8$ &  DeepReDuce \citep{jha2021deepreduce} & $88.9$ & $2.41$ & $8\times$\\
			\hline
			& VGG16 & $93.8$ & $126$ & SENet(ours)  & $93.42$ & $0.74$ & $2.3\times$\\
			\cline{2-4} \cline{6-8}
			 & ResNet18 & 95.2 & $150$ &   & $\bm{94.91}$ & $0.63$ & $3.7\times$\\
			\cline{2-8}
			 $ 100$k $\le r \le 500 $k & VGG16 & $93.8$ & $126$ &  DeepReDuce \citep{jha2021deepreduce} & $92.5$ & $0.73$ & $2.3\times$\\
			\cline{2-8}
			 & VGG16 & $93.8$ & $126$ &  SAFENet \citep{lou2020safenet} & $88.9$ & $0.7$ & $2.3\times$ \\
			\cline{2-8}
			 & Custom Net & $95.0$ & $100$ &  CryptoNAS \citep{ghodsi2020cryptonas} & $92.18$ & $0.92$ & --\\
			 &  &  & $500$ &  & $94.41$ & $0.19$ & --\\
			\hline
			\hline
			 \multicolumn{8}{c}{Dataset: CIFAR-100}\\
			\hline
			& ResNet18 & $78.05$ & $24.6$ &   & $70.59$ & $2.87$ & $21.8\times$ \\
			 &  &  & $49.6$ &  & $75.28$ & $1.52$ & $11.2\times$ \\
			&  &  & $100$ & SENet(ours)   & $\bm{77.92}$ & $0.78$ & $5.6\times$ \\
		    \cline{2-4} \cline{6-8}
		    $ 0 \le r \le 100 $k & ResNet34 & $78.42$ & $50.1$ &   & $74.84$ & $1.5$ & $19.3\times$\\
			 &  &  & $80$ &   & $76.66$ & $0.96$ & $12.1\times$\\
			\cline{2-8}
			 & ResNet18 & $78.05$ & $28.7$ &  DeepReDuce \citep{jha2021deepreduce} & $68.6$ & $2.39$ & $19.4\times$\\
			 &  &  & $49.2$ &  & $69.5$ & $1.41$ &$11.3\times$ \\
			\cline{2-8}
			 & Custom Net & $74.93$ & $51$ &  Sphynx \citep{cho2021sphynx} & $69.57$ & $1.36$ & --\\
			\hline
			& ResNet18 & $78.05$ & $150$ &   & $78.32$ & $0.52$ & $3.7\times$ \\
			\cline{2-4} \cline{6-8}
			& ResNet34 & $78.425$ & $200$ &   & $78.8$ & $0.4$ & $4.8\times$\\
			\cline{2-4} \cline{6-8}
			& \textcolor{black}{WRN22-8} & \textcolor{black}{$80.82$} & \textcolor{black}{$180$} & SENet(ours)  & \textcolor{black}{$79.12$} & \textcolor{black}{$0.44$} & \textcolor{black}{$7.7\times$} \\
			&  &  & $240$ &   & $79.81$ & $0.33$ & $5.8\times$ \\
			&  &  & $300$ &   & $\bm{80.54}$ & $0.27$ & $4.6\times$ \\
			\cline{2-8}
			 & \textcolor{black}{WRN22-8} & \textcolor{black}{$80.82$} & \textcolor{black}{$180$} &  \textcolor{black}{SNL} \citep{cho2022selective} & \textcolor{black}{$77.65$} & \textcolor{black}{$0.43$} & \textcolor{black}{$7.7\times$}\\
			\cline{2-8}
			$ 100$k $ \le r \le 500 $k & ResNet18 & $78.05$ & $229.4$ &  DeepReDuce \citep{jha2021deepreduce} & $76.22$ & $0.33$ & $2.4\times$\\
			\cline{2-8}
			& Custom Net & $74.93$ & $102$ &  Sphynx \citep{cho2021sphynx} & $72.9$ & $0.714$ & --\\
			&   &  & $230$ &   & $74.93$ & $0.32$ & --\\
			\cline{2-8}
			 & Custom Net & $79.07$ & $100$ &  CryptoNAS \citep{ghodsi2020cryptonas} & $68.67$ & $0.69$ & --\\
			 &  &  & $500$ &  & $77.69$ & $0.16$ & --\\
		\hline
		\end{tabular}
\end{table} 

\textbf{Results with activation channel level ReLU reduction.} As shown in Table \ref{table:channel_act_relu}, while trimming ReLUs at a higher granularity of activation channel level, SENet models suffer a little more drop in accuracy compared to that at pixel level. For example, at a ReLU budget of 240k, channel-level ReLU removal yields an accuracy of $79.3\%$ compared to $79.81\%$ of pixel-level. However, compared to existing alternatives, SENet can achieve improved performance of up to $1.85\%$ for similar ReLUs.  

\subsection{SENet++ Results}
\begin{figure}[!t]
\vspace{-5mm}
  \begin{center}
    \includegraphics[width=0.86\textwidth]{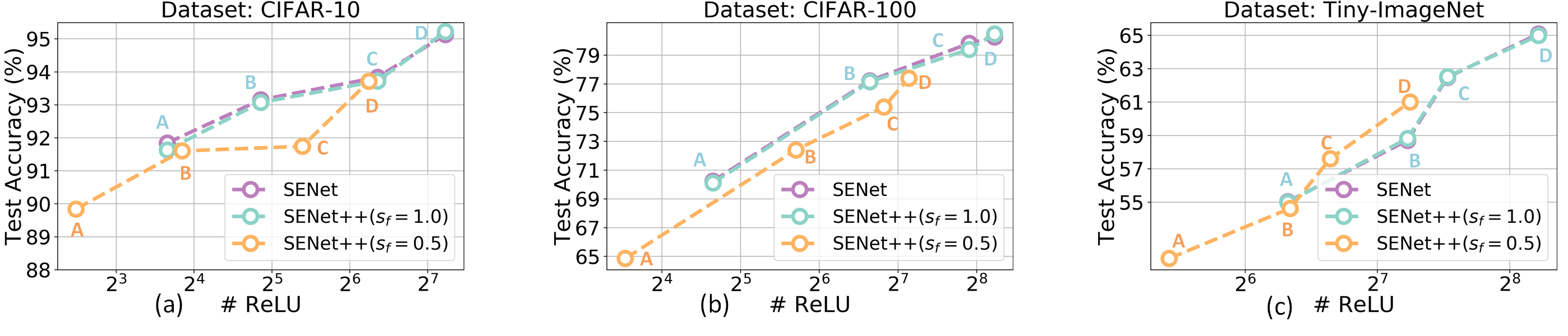}
  \end{center}
  \vspace{-4mm}
  \caption{Performance of SENet++ on three datasets for various $\#$ReLU budgets. The points labeled A, B, C, and D corresponds to experiments of different target \#ReLUs for the full model ($d_r = 1.0$). For SENet++, note that a single training loop yields two points with the same label corresponding to the two different dropout rates.}
  \label{fig:senet_vs_senet_pp}
  \vspace{-2mm}
\end{figure}

For SENet++, we performed experiments with $\mathcal{D}_r = [0.5, 1.0]$, meaning each training loop can yield models with two different channel dropout rates. The $0.5$-sub-model enjoys a $\mathord{\sim}4\times$ MACs reduction compared to the full model. Moreover, as shown in Fig. \ref{fig:senet_vs_senet_pp}, the $0.5$-sub-model also requires significantly less $\#$ReLUs due to reduced model size. In particular, the smaller models have $\#$ReLUs reduced by a factor of $2.05\times$, $2.08\times$, and $1.88\times$ on CIFAR-10, CIFAR-100, and Tiny-ImageNet, respectively, compared to the PR full models, averaged over four experiments with different ReLU budgets for each dataset. \textit{Lastly, the similar performance of the SENet and SENet++ models at $d_r  = 1.0$ with similar ReLU budgets, clearly depicts the ability of SENet++ to yield multiple sub-models without sacrificing any accuracy for the full model}.

\subsection{Analysis of Linear and ReLU Inference Latency}
\begin{figure}[!t]
\vspace{0mm}
  \begin{center}
    \includegraphics[width=0.86\textwidth]{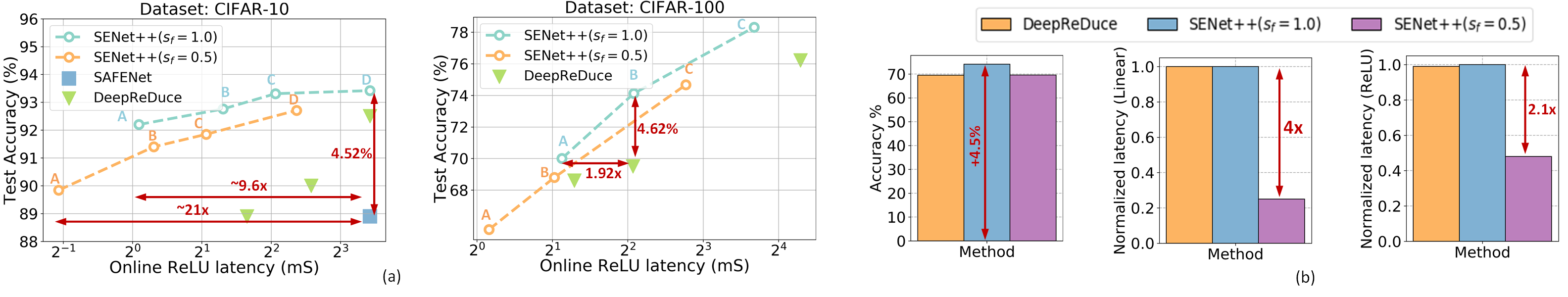}
  \end{center}
  \vspace{-5mm}
  \caption{Performance comparison of SENet++ (with $d_r = 1.0$ and $0.5$) vs. existing alternatives (a) with VGG16 and ResNet18 in terms of ReLU latency. The labels A, B, C, D correspond to experiments of different target \#ReLUs for the full model ($d_r = 1.0$). For SENet++, note that a single training loop yields two points with the same label corresponding to the two different dropout rates. (b) Comparison between DeepReDuce and SENet++ for a target $\#$ ReLU budget of $\mathord{\sim}50$k with ResNet18 on CIFAR-100. }
  \label{fig:vgg16_c10_res18_c100_comparison}
  \vspace{-5mm}
\end{figure}

Table \ref{table:cost_report} shows the GC-based online ReLU operation latency is $\mathord{\sim}343\times$ higher than one linear operation (multiply and accumulate), making the ReLU operation latency the dominant latency component. Inspired by this observation, we quantify the online PI latency as that of the $N$ ReLU operations for a model with ReLU budget of $N$. In particular, based on this evaluation, Fig. \ref{fig:vgg16_c10_res18_c100_comparison}(a) shows the superiority of SENet++ of up to $\mathord{\sim}9.6\times$ ($\mathord{\sim}1.92\times$) reduced online ReLU latency on CIFAR-10 (CIFAR-100). With negligibly less accuracy this latency improvement can be up to $\mathord{\sim}21\times$. Furthermore, when $d_r < 1.0$, SENet++ requires fewer MACs and the linear operation latency can be significantly reduced, as demonstrated in Fig. \ref{fig:vgg16_c10_res18_c100_comparison}(b).
\subsection{Ablation Studies}

\textbf{Importance of ReLU sensitivity.}
\textcolor{black}{To understand the importance of layer-wise ReLU sensitivity evaluations at a given ReLU budget, we conducted experiments with evenly allocated}
\begin{wraptable}{r}{0.43\textwidth}
\vspace{-5mm}
	\tiny\addtolength{\tabcolsep}{-5.3pt}
		\caption{Importance of ReLU sensitivity.}
		\label{table:abl_importance_relu_sens}
		\centering
		\begin{tabular}{c|c|c|c|c|c|c}
			\hline
			 \textbf{Model} & \textbf{Baseline} & \#\textbf{ReLU}  & \textbf{ReLU} & \textbf{Test} & \textbf{Acc}\%/ & \textbf{Comm.}\\
			 & \textbf{Acc}\% &  \textbf{(k)}  & \textbf{Sensitivity} &\textbf{Acc}\% & \#\textbf{1k ReLU} &  \textbf{Savings} \\
			 \hline
			  &  & $139.2$ & \xmark & $70.12$ & $0.503$ & $\times 4$ \\
			  ResNet18 & $78.05$ &   $135$  & \cmark & $\bm {75.88}$ & $0.56$ & $\times 4.12$ \\
			  & &   $70.4$  & \cmark & $73.03$ & $1.03$ & $\times \bm{7.9}$ \\
			\hline
		\end{tabular}
		\vspace{-3mm}
\end{wraptable}
ReLUs. 
Specifically, for ResNet18, for a ReLU budget of $25\%$ \textcolor{black}{of that of the}
original model, we randomly 
removed $75\%$ ReLUs from each PR layer with identity elements to create the ReLU mask, and trained the PR model with this mask. We further trained two other PR
 ResNet18 with similar and lower \# ReLU budgets with the per-layer ReLUs assigned following the proposed sensitivity. As shown in Table \ref{table:abl_importance_relu_sens}, the sensitivity-driven PR models can yield significantly improved performance of $\mathord{\sim}5.76\%$ for similar ReLU budget, demonstrating the importance of proposed ReLU sensitivity.

\textbf{Choice of the hyperparameter $\lambda$ and $\beta$.}
To determine the influence of the AR teacher's influence
\begin{wrapfigure}{r}{0.47\textwidth}
\vspace{-5mm}
  \begin{center}
    \includegraphics[width=0.468\textwidth]{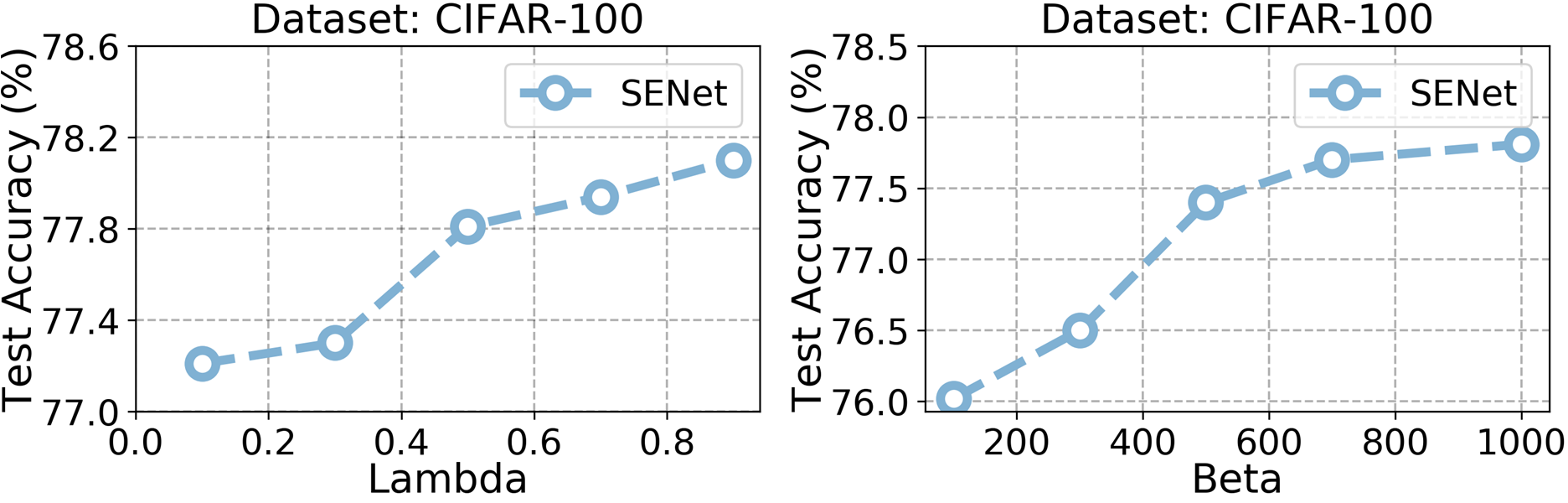}
  \end{center}
  \vspace{-4mm}
  \caption{Ablation studies with different $\lambda$ and $\beta$ values for the loss term in Eq. \ref{eq:sp_distill_loss}.}
  \vspace{-5mm}
  \label{fig:ablation_lambda_beta}
\end{wrapfigure}
 on the PR model's learning, we conducted the final stage distillation with $\lambda \in [0.1, 0.3, 0.5, 0.7, 0.9]$ and $
\beta \in [100, 300, 500, 700, 1000]$. As shown in Fig. \ref{fig:ablation_lambda_beta}, the performance of the student PR model improves with the increasing influence of the teacher both in terms of high $\lambda$ and $\beta$ values. However, we also observe, the performance improvement \textcolor{black}{tends} to saturate at $\beta \approx 1000$. \textcolor{black}{Note, we keep $\lambda=0.5$ and $\beta=1000$, for the $\beta$ and $\lambda$ 
\textcolor{black}{ablations}, respectively.}

\section{Conclusions}
In this paper, we introduced the notion of ReLU sensitivity for non-linear layers of a DNN model. Based on this notion, we present an automated ReLU allocation and training algorithm for models with limited ReLU budgets that targets latency and communication-efficient PI. The resulting networks can achieve similar to SOTA accuracy while significantly reducing the $\#$ ReLUs by up to $9.6\times$ on CIFAR-10, enabling a dramatic reduction of the latency and communication costs of PI. \textcolor{black}{Extending this idea of efficient PI to vision transformer models is an interesting future research.}

\section{Acknowledgment}
This work was supported in parts by Intel, DARPA
under the agreement numbers HR00112190120, and FA8650-18-1-7817.

\bibliography{iclr2023_conference}
\bibliographystyle{iclr2023_conference}

\appendix
\section{Appendix}
\subsection{Training Hyperparameters and Models}
The training hyperparameter details for each stage is provided in table \ref{table:supp_hyper_details}. Also, for the KL-divergence loss in stage 2 and 3, we fix the temperature value $\rho =4.0$. To evaluate $\eta_{\bm {\theta}}$, we a set of $1000$ randomly selected training image samples.

Table \ref{table:supp_senet_pp_loss} details the loss function used for different OD rate values for the SENet++ (at the fine-tuning stage of the PR model i.e. training stage 3). Note, the training stage 1 where the AR model gets trained, we use standard CE loss for all the sub-models with different OD rate values.
The detailed fine-tune training algorithm is provided in \ref{alg:senet_pp_stage3}. Note, for SENet, the algorithm remains the same with $\mathcal{D}_r = [1.0]$, meaning supporting only the full model with all the channels present\footnote{We have open-sourced the validation codes with the supplementary.}.

\textcolor{black}{\textbf{Model Selection}. For lower resolution images (CIFAR-10, CIFAR-100, Tiny-ImageNet) compared to that of ImageNet, we have used the variant of ResNet18 and ResNet34 models that are suitable for supporting lower resolution datasets. In particular, we replaced the $7\times7$ kernel, stride 2, and padding 3 of the first layer with a $3\times3$ kernel having a stride and padding of 1 each. We would also like to highlight that this is a popular practice and can be seen in various other peer-reviewed manuscripts \citep{wang2020once,liu2020autocompress,wong2020fast}. More importantly, both the existing state-of-the-art methods \citep{cho2022selective,jha2021deepreduce} used similar ResNet models as ours, for their evaluations at a reduced ReLU budget.}

\begin{table}[!h]
	\scriptsize\addtolength{\tabcolsep}{-2.5pt}
		\caption{Hyperparameter settings of SENet/SENet++ training method.}
		\label{table:supp_hyper_details}
		\centering
		\begin{tabular}{c|c|c|c|c|c|c|c|c|c|c|c}
			\hline
			\textbf{Model(s)} & \textbf{Dataset} &  \multicolumn{3}{c|}{\textbf{Epoch}} & \textbf{batch}  & \multicolumn{3}{c|}{\textbf{Initial LR}} & \textbf{Momen-} & \textbf{Optim-} & \textbf{Weight} \\
			\cline{3-5} \cline{7-9}
			& & \textbf{stage1} & \textbf{stage2} & \textbf{stage3} & \textbf{-size} & \textbf{stage1} & \textbf{stage2} & \textbf{stage3} & \textbf{tum} & \textbf{izer} & \textbf{decay} \\ 
			 \hline
			ResNet18, & CIFAR-10 & 240 & 150 & 180 & 128 & 0.05 & 0.05 & 0.01 & 0.9 & SGD & 0.0005 \\
			VGG16 & & & & & & & & & & & \\
			\hline
			ResNet\{18, 34\} & CIFAR-100 & 240 & 150 & 180 & 128 & 0.05 & 0.05 & 0.01 & 0.9 & SGD & 0.0005 \\
			WRN22-8 & & & & & & & & & & & \\
			\hline
			ResNet18 & Tiny-ImageNet & 120 & 100 & 120 & 32 & 0.05 & 0.05 & 0.01 & 0.9 & SGD & 0.0005 \\
			\hline
			ResNet18 & ImageNet & 60 & 30 & 50 & 16 & 0.05 & 0.05 & 0.01 & 0.9 & SGD & 0.0005 \\
		\hline
		\end{tabular}
\end{table} 
\begin{table}[!h]
	\scriptsize\addtolength{\tabcolsep}{-0.0pt}
		\caption{Loss function used for sub-models with different OD rates in SENet++ at training stage 3.}
		\label{table:supp_senet_pp_loss}
		\centering
		\begin{tabular}{c|c}
			\hline
			\textbf{OD rate ($d_r$)} & \textbf{Loss} \\
			\hline
			$1.0$ & (1-$\lambda$)$\mathcal{L}_{CE}$ + $\lambda$ $\mathcal{L}_{KL}$ +$\frac{\beta}{2}$ $\mathcal{L}_{PRAM}$ \\ \hline
			$< 1.0$ & (1-$\lambda$)$\mathcal{L}_{CE}$ + $\lambda$ $\mathcal{L}_{KL}$ \\
		\hline
		\end{tabular}
\end{table}

\begin{algorithm}[!h]
\small
\SetAlgoLined
\DontPrintSemicolon
\KwData{Trained AR model parameters $\bm{\Theta}_{AR}$, PR model parameters $\bm{\Theta}_{PR}$, mini-batch size $\mathcal{B}$, learned ReLU mask from stage 2 $\mathbf{\Pi}$, OD set $\mathcal{D}_r$.},
\textbf{Output:} Trained PR model with \# ReLU count $r$.\\
$ \bm{\Theta}_{PR} \leftarrow \texttt{applyModelWeight(}\bm{\Theta}_{AR},{\mathbf{\Pi}} \text{)} $\;

\For{$\text{i} \leftarrow 0$ \KwTo \KwTo {$ep$}}
{
   
    \For{$\text{j} \leftarrow 0$ \KwTo {${n_{\mathcal{B}}}$}}
    {
     \For{$\text{dropout rate}$ $d_r$ in sorted $\mathcal{D}_r$}
     {
        $\texttt{sampleSBN}(d_r)\text{ //sample the BN } \text{ corresponding to } d_r$\;
        $\mathcal{L}_{CE} \leftarrow \texttt{computeCELoss(}\bm{X}_{0:{\mathcal{B}}}$, $\bm{Y}_{0:{\mathcal{B}}}, \bm{Y}^{pr}_{0:{\mathcal{B}}}, {\mathbf{\Pi}})$\;
        $\mathcal{L}_{KL} \leftarrow \texttt{computeKLLoss(}f_{\Phi}(\Theta_{PR}), f_{\Phi}(\Theta_{AR}), \rho, {\mathbf{\Pi}})$\;
        \eIf{$d_r == 1.0$}
        {
        $\mathcal{L}_{PRAM} \leftarrow \texttt{computePRAMLoss(}f_{\Phi}(\Theta_{PR}), f_{\Phi}(\Theta_{AR}), {\mathbf{\Pi}})$\;
        $\mathcal{L} \leftarrow (1-\lambda)\mathcal{L}_{CE} + \lambda \mathcal{L}_{KL} + \frac{\beta}{2} \mathcal{L}_{PRAM}$\;
        }
        {
        $\mathcal{L} \leftarrow (1-\lambda)\mathcal{L}_{CE} + \lambda \mathcal{L}_{KL}$\;
      }
      $\texttt{accumulateGrad}(\mathcal{L})$\;
      }  
        $\texttt{updateParam}(\bm{\Theta}_{PR}, \nabla_{\mathcal{L}})$\;
    }
}
 \caption{SENet++ Fine-Tune Training Stage}
 \label{alg:senet_pp_stage3}
\end{algorithm}

\subsection{\textcolor{black}{Model Latency Estimation}}
\textcolor{black}{Similar to the existing literature \citep{jha2021deepreduce, cho2022selective, lou2020safenet}, we assume the popular PI framework described in Delphi \citep{mishra2020delphi}, and leverage their reported per ReLU operation latency to estimate the total online cryptographic private inference latency. In particular, assuming sequential execution, the total latency can be estimated as}
\begin{align}
\textcolor{black}{T = (N_{MAC} * t_{mac} + N_{ReLU} * t_{relu}) \mu S}. 
\end{align}
\noindent
\textcolor{black}{where $N_{MAC}$ and $N_{ReLU}$ corresponds to the total number of MAC and ReLU operations required to perform a single forward pass.
$t_{relu}$ represents per ReLU cipher-text execution time and its value is assumed as $85.3\mu$S. Note, Per ReLU cipher text execution wall-clock time is taken from \citep{mishra2020delphi} and the execution of 1000 ReLUs takes proportional time as reported in \citep{cho2022selective}. Thus similar to \citep{cho2022selective, jha2021deepreduce}, we extrapolated per-ReLU wall clock time to extract total ReLU latency for cryptographic inference. Similarly, we evaluated the per linear operation latency $t_{mac}$ to be $0.248\mu$S \citep{mishra2020delphi}.
Now, as cryptographic ReLU latency is around $343\times$  costlier than that for linear ops, similar to earlier literature \citep{jha2021deepreduce, cho2022selective, lou2020safenet}, for comparable $N_{MAC}$ and $N_{ReLU}$ we can approximate the total latency with that of the non-linear latency.} Note, the values of the ReLU and linear operation latency are taken from \citep{mishra2020delphi}, however, we understand with recent improvement of operation latency the `per-operation delay' can reduce \citep{huang2022cheetah}.  

\subsection{\textcolor{black}{Discussion on the relation between ReLU importance and pruning sensitivity}}
\begin{figure}[!ht]
\vspace{-4mm}
  \begin{center}
    \includegraphics[width=0.60\textwidth]{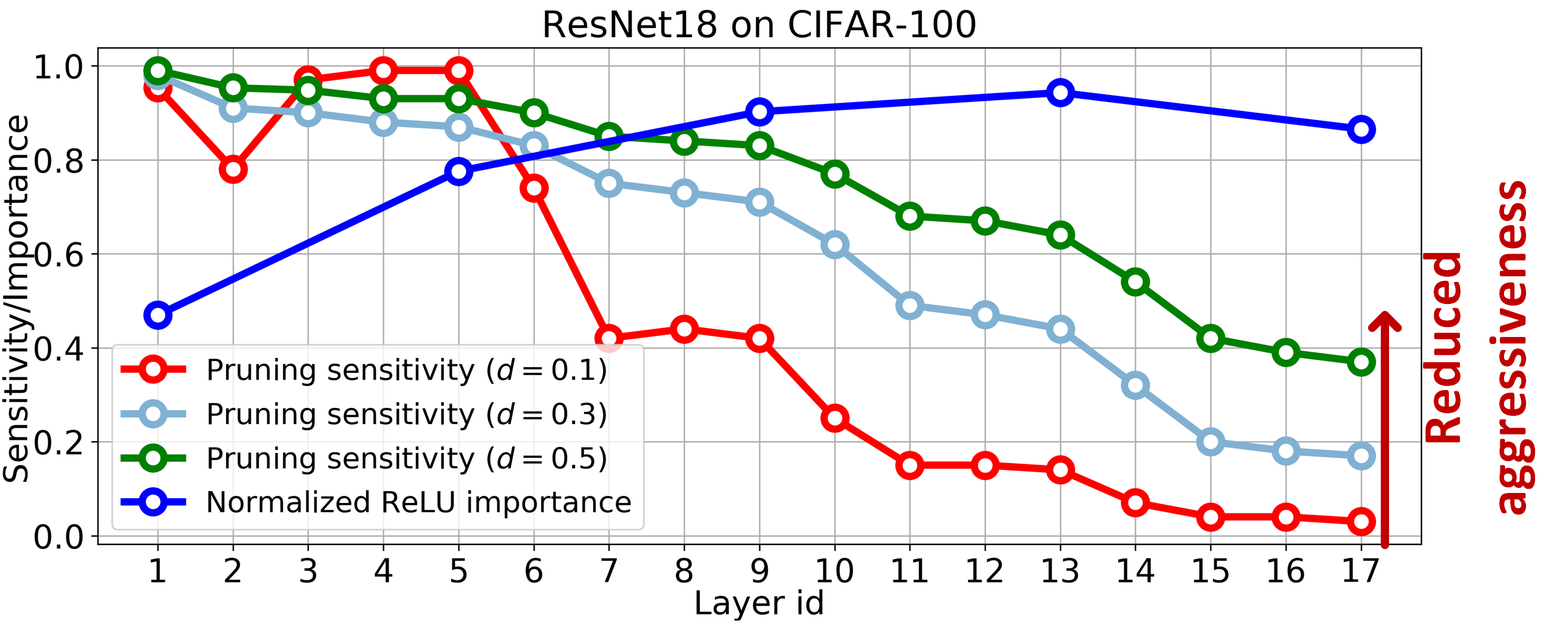}
  \end{center}
    \vspace{-4mm}
  \caption{Ablation with different $d$ yielding different layer-wise parameter pruning sensitivity. It can be clearly observed that as the $d$ increases, the aggressiveness of sensitivity change reduces and becomes similar to what we observe for the ReLU importance plot.}
  \label{fig:importance_vs_sens_abl}
  \vspace{-1mm}
\end{figure}

\textcolor{black}{Fig. \ref{fig:relu_utility_prune_sens} of the main manuscript demonstrated the inverse trend between ReLU importance and parameter pruning sensitivity. Here, as an exemplary, we used an extremely low target parameter density $d=0.1$ to compute the parameter sensitivity. This choice forces an aggressive drop in pruning sensitivity of later layers because they correspond to the majority of the model parameters. In contrast, when we set a less aggressive compression with higher parameter density $d$  of 0.3 or 0.5, we observe a more gradual reduction in weight pruning sensitivity at later layers similar to change observed with ReLU importance. More importantly, as shown in Fig. \ref{fig:importance_vs_sens_abl}, the inverse trend between ReLU importance and pruning sensitivity can still be clearly observed irrespective of the choice of $d$, while abruptness of sensitivity change remains a function of target $d$.}

\subsection{More results and Analysis}

\begin{figure}[!ht]
\vspace{-3mm}
  \begin{center}
    \includegraphics[width=0.60\textwidth]{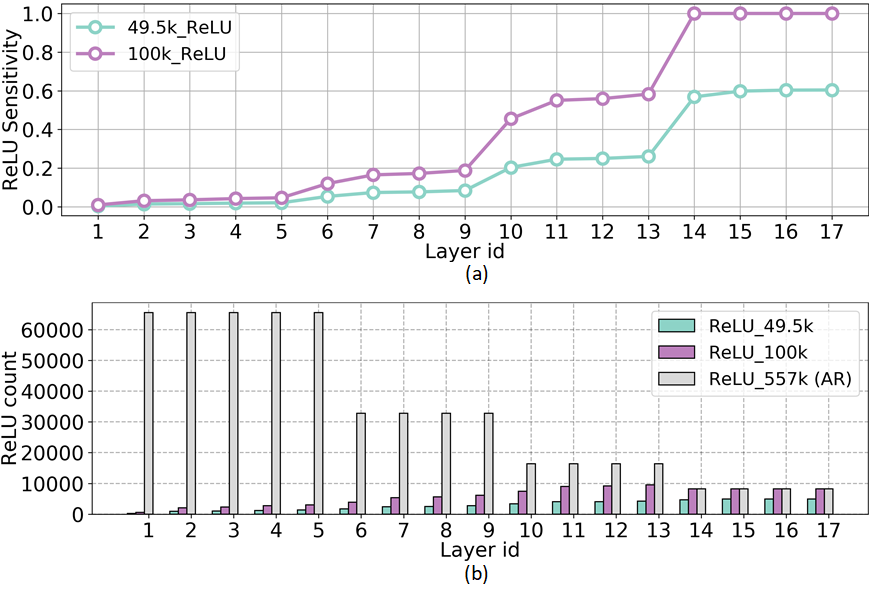}
  \end{center}
  \vspace{-4mm}
  \caption{(a) ReLU sensitivity per non-layer layer meeting different target ReLU budget, (b) ReLU count for different layers, evaluated following the sensitivity. We used ResNet18 on CIFAR-100 for this analysis.}
  \label{fig:supp_relu_count_sens}
  \vspace{-5mm}
\end{figure}  

\textbf{Layer-wise ReLU Sensitivity.} Fig. \ref{fig:supp_relu_count_sens}(a) shows the ReLU sensitivity per layer. As elaborated in the Section 3, the ReLU sensitivity is more in the later layers, making our ReLU sensitivity follow similar trend as ReLU importance in DeepReDuce \citep{jha2021deepreduce}. Fig. \ref{fig:supp_relu_count_sens}(b) shows the layer wise ReLU count for different ReLU budget, allocation driven by sensitivity (Fig. \ref{fig:supp_relu_count_sens}(a)). With the increasing count of ReLU budget, the assignment of ReLU happens more aggressively at the later layers, compared to the earlier ones. 

\begin{figure}[!ht]
\vspace{-0mm}
  \begin{center}
    \includegraphics[width=0.99\textwidth]{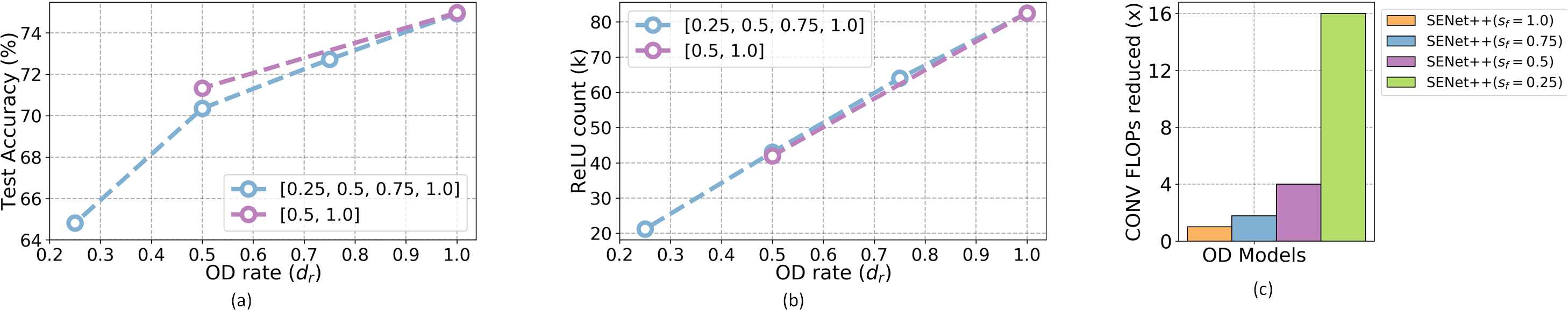}
  \end{center}
  \caption{(a) Accuracy vs. OD rate, (b) ReLU count vs. OD rate for SENet++ training with different OD rate supports (\textit{two} and \textit{four}), (c) CONV layer FLOPs reduction factor for models at different OD rate values. We used ResNet18 on CIFAR-100 for this evaluation.}
  \label{fig:supp_perf_at_OD_four_values}
  \vspace{-0mm}
\end{figure}

\begin{figure}[!ht]
\vspace{-0mm}
  \begin{center}
    \includegraphics[width=0.80\textwidth]{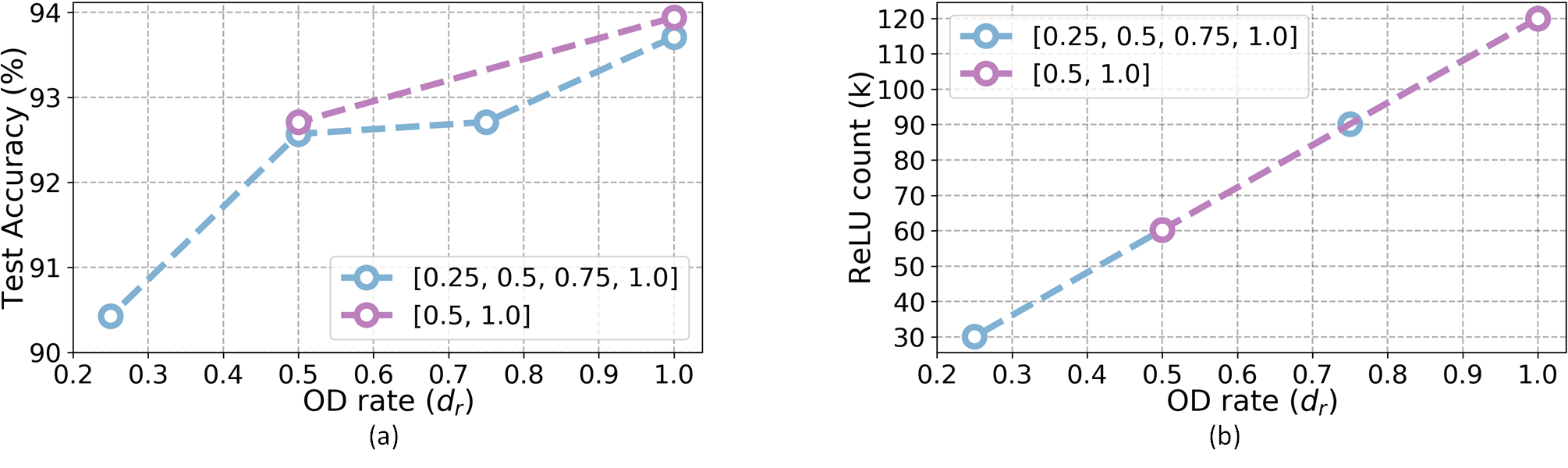}
  \end{center}
  \caption{(a) Accuracy vs. OD rate, (b) ReLU count vs. OD rate for SENet++ training with different OD rate supports (\textit{two} and \textit{four}). We used VGG16 on CIFAR-10 for this evaluation.}
  \label{fig:supp_perf_at_OD_four_vgg16_values}
  \vspace{-0mm}
\end{figure}

\textbf{SENet++: Training with More Than 2 Dropout Rates.}
In the original manuscript we showed results with two dropout rates ($\mathcal{D}_r = [0.5, 1.0]$), while training with different dropouts, and yield models with different channel width factors. We now show results with $\mathcal{D}_r = [0.25, 0.5, 0.75, 1.0]$. In particular, Fig. \ref{fig:supp_perf_at_OD_four_values} and \ref{fig:supp_perf_at_OD_four_vgg16_values} show the results with models yielded via training on \textit{four} different $d_r$ choices. As shown in the Fig. \ref{fig:supp_perf_at_OD_four_values}(a) and \ref{fig:supp_perf_at_OD_four_vgg16_values}(a), the performance of the models at $d_r=1.0$ and $d_r=0.5$, are similar for both two and four $d_r$ choices, making SENet++ an efficient algorithm in yielding multiple reduced FLOPs/ReLU models. Moreover, the effective ReLU reduction remains proportional to the corresponding ordered dropout rate (OD rate) (Fig. \ref{fig:supp_perf_at_OD_four_values}(b) and \ref{fig:supp_perf_at_OD_four_vgg16_values}(b)). Fig.\ref{fig:supp_perf_at_OD_four_values}(c) shows the effective FLOPs reduction for the CONV layers while performing inference at a reduced OD rate model selection. 

\subsection{\textcolor{black}{Further Ablation studies}}
\textbf{\textcolor{black}{Ablation Studies for Stage 2 Importance.}}
\textcolor{black}{To understand the importance of the ReLU mask identification stage, we now present results for SENet with and without that stage. In particular, for the model without stage 2, we randomly assign the ReLU mask by following the ReLU layer sensitivity, meaning layers having higher sensitivity will have non-zero mask values of similar proportions at random locations of the corresponding ReLU mask tensors.Table \ref{table:abl_imp_stage2} clearly demonstraates the benefits of stage 2 as the model can provide improved performance of 11.28\% compared to the one trained without stage 2.}

\begin{table}[!h]
\vspace{-5mm}
	\small\addtolength{\tabcolsep}{-0pt}
		\caption{\textcolor{black}{Importance of ReLU mask identification stage (stage 2). We used CIFAR-100 dataset.}}
		\label{table:abl_imp_stage2}
		\centering
		\begin{tabular}{c|c|c|c|c|c}
			\hline
			 \textbf{Model} & \textbf{Baseline} & \#\textbf{ReLU}  & \textbf{ReLU mask} & \textbf{Test} & \textbf{Acc}\%/ \\
			 & \textbf{Acc}\% &  \textbf{(k)}  & \textbf{identification (stage 2)} &\textbf{Acc}\% & \#\textbf{1k ReLU} \\
			 \hline
			  &  & $24.6$ & \xmark & $59.12$ & $2.37$ \\
			  ResNet18 & $78.05$ &   $24.6$  & \cmark & $\bm {70.59}$ & $2.87$ \\
			\hline
		\end{tabular}
		\vspace{-0mm}
\end{table}

\textbf{\textcolor{black}{Ablation Studies for Stage 3 Importance.}}
\textcolor{black}{Table \ref{table:abl_imp_stage3} shows the importance of fine-tuning stage (stage 3). In particular, the accuracy difference of a model before and after stage 3 training can vary up to 6.3\%.}

\begin{table}[!h]
\vspace{-5mm}
	\small\addtolength{\tabcolsep}{-0pt}
		\caption{\textcolor{black}{Importance of Activation similarity maximization (stage 3). We used CIFAR-100 dataset.}}
		\label{table:abl_imp_stage3}
		\centering
		\begin{tabular}{c|c|c|c|c|c}
			\hline
			 \textbf{Model} & \textbf{Baseline} & \#\textbf{ReLU}  & \textbf{Activation similarity} & \textbf{Test} & \textbf{Acc}\%/ \\
			 & \textbf{Acc}\% &  \textbf{(k)}  & \textbf{maximization (stage 3)} &\textbf{Acc}\% & \#\textbf{1k ReLU} \\
			 \hline
			  &  & $24.6$ & \xmark & $64.10$ & $2.57$ \\
			  ResNet18 & $78.05$ &   $24.6$  & \cmark & $\bm {70.59}$ & $2.87$ \\
			\hline
		\end{tabular}
		\vspace{-0mm}
\end{table}

\textbf{\textcolor{black}{Ablation Studies with the PRAM Loss Component.}}
\textcolor{black}{Table \ref{table:abl_pram} shows the importance of PRAM loss component during the fine-tuning stage (stage 3). In particular, the accuracy can improve up to 0.69\% on CIFAR-10 as evaluated with ResNet18 for 150k ReLU budget.}

\begin{table}[!h]
\vspace{-5mm}
	\small\addtolength{\tabcolsep}{-0pt}
		\caption{\textcolor{black}{Importance of PRAM loss at final fine-tuning stage. We used CIFAR-10 dataset.}}
		\label{table:abl_pram}
		\centering
		\begin{tabular}{c|c|c|c|c|c}
			\hline
			 \textbf{Model} & \textbf{Baseline} & \#\textbf{ReLU}  & \textbf{With PRAM} & \textbf{Test} & \textbf{Acc}\%/ \\
			 & \textbf{Acc}\% &  \textbf{(k)}  & \textbf{loss} &\textbf{Acc}\% & \#\textbf{1k ReLU} \\
			 \hline
			  &  & $150$ & \xmark & $94.12$ & $0.62$ \\
			  ResNet18 & $95.2$ &   $150$  & \cmark & $\bm {94.91}$ & $0.63$ \\
			\hline
		\end{tabular}
		\vspace{-0mm}
\end{table}

\end{document}